%% file: neurips_2024.tex
\documentclass{article}

\usepackage[numbers]{natbib}
\usepackage[final]{neurips_2024}

\usepackage[utf8]{inputenc} 
\usepackage[T1]{fontenc}    
\usepackage{hyperref}       
\usepackage{url}            
\usepackage{booktabs}       
\usepackage{amsfonts}       
\usepackage{nicefrac}       
\usepackage{microtype}      
\usepackage{xcolor}         
\usepackage{tabularx}       
\usepackage{graphicx}
\usepackage{lipsum}
\usepackage{marvosym}
\usepackage{xspace}
\usepackage{blindtext}
\usepackage{amsmath}
\usepackage{algorithm}
\usepackage{algpseudocode}
\usepackage{wrapfig}
\usepackage{enumitem}

\makeatletter
\DeclareRobustCommand\onedot{\futurelet\@let@token\@onedot}
\def\@onedot{\ifx\@let@token.\else.\null\fi\xspace}

\def\etc{\emph{etc}\onedot}

\makeatother

\def\shortname{\mbox{Hunyuan3D 2.0}\xspace}

\definecolor{citecolor}{HTML}{0037a6}
\definecolor{linkcolor}{HTML}{ED1C24}
\definecolor{graycolor}{rgb}{0.95,0.95,0.95}

\usepackage[capitalize]{cleveref}
\crefname{section}{Sec.}{Secs.}
\crefname{table}{Tab.}{Tabs.}
\crefname{figure}{Fig.}{Figs.}

\title{
\shortname: Scaling Diffusion Models for High Resolution Textured 3D Assets Generation
}

\author{
}
\begin{document}

\maketitle

\vspace{-20mm} 
\begin{quote}
    \quad ``\textit{
    Living out everyone's imagination on creating and manipulating 3D assets.}''\flushright{ --- \textbf{Hunyuan3D Team $^*$} }
\end{quote}
\vspace{5mm}

\input{sections/abstract}

\begin{figure}[h]
\centering
\includegraphics[width=0.82\textwidth]{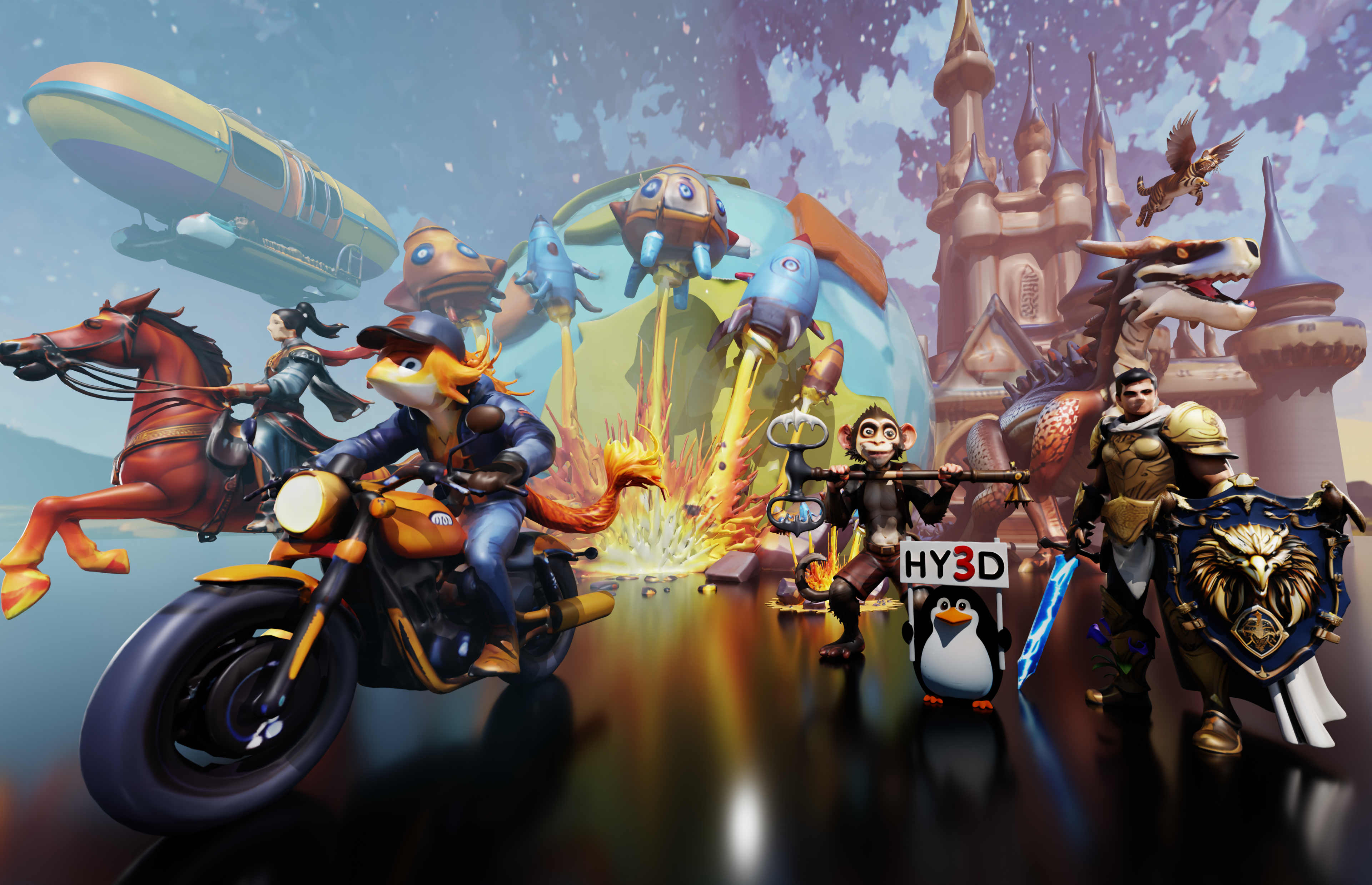}

\end{figure}

\makeatletter
\def\@makefnmark{}
\makeatother
\footnotetext{$*$ Hunyuan3D team contributors are listed in the end of report.}

\clearpage

\input{sections/system}

\clearpage

\input{sections/introduction}
\input{sections/method}

\input{sections/experiments}

\clearpage

\input{sections/applications}
\input{sections/related-work}

\input{sections/conclusion}

\clearpage

{\small
\bibliographystyle{plain}
\bibliography{references}
}

\end{document}

%% file: sections/abstract.tex
\begin{abstract}
We present \shortname, an advanced large-scale 3D synthesis system for generating high-resolution textured 3D assets. This system includes two foundation components: a large-scale shape generation model -- Hunyuan3D-DiT, and a large-scale texture synthesis model -- Hunyuan3D-Paint. 
The shape generative model, built on a scalable flow-based diffusion transformer, aims to create geometry that properly aligns with a given condition image, laying a solid foundation for downstream applications. 
The texture synthesis model, benefiting from strong geometric and diffusion priors, produces high-resolution and vibrant texture maps for either generated or hand-crafted meshes.
Furthermore, we build Hunyuan3D-Studio -- a versatile, user-friendly production platform that simplifies the re-creation process of 3D assets. It allows both professional and amateur users to manipulate or even animate their meshes efficiently.
We systematically evaluate our models, showing that \shortname outperforms previous state-of-the-art models, including the open-source models and closed-source models in geometry details, condition alignment, texture quality, and \etc.
\shortname is publicly released in order to fill the gaps in the open-source 3D community for large-scale foundation generative models.
The code and pre-trained weights of our models are available at:
\url{https://github.com/Tencent/Hunyuan3D-2}.
\end{abstract}

%% file: sections/system.tex


\begin{figure}[t]
\centering
\includegraphics[width=1\textwidth]{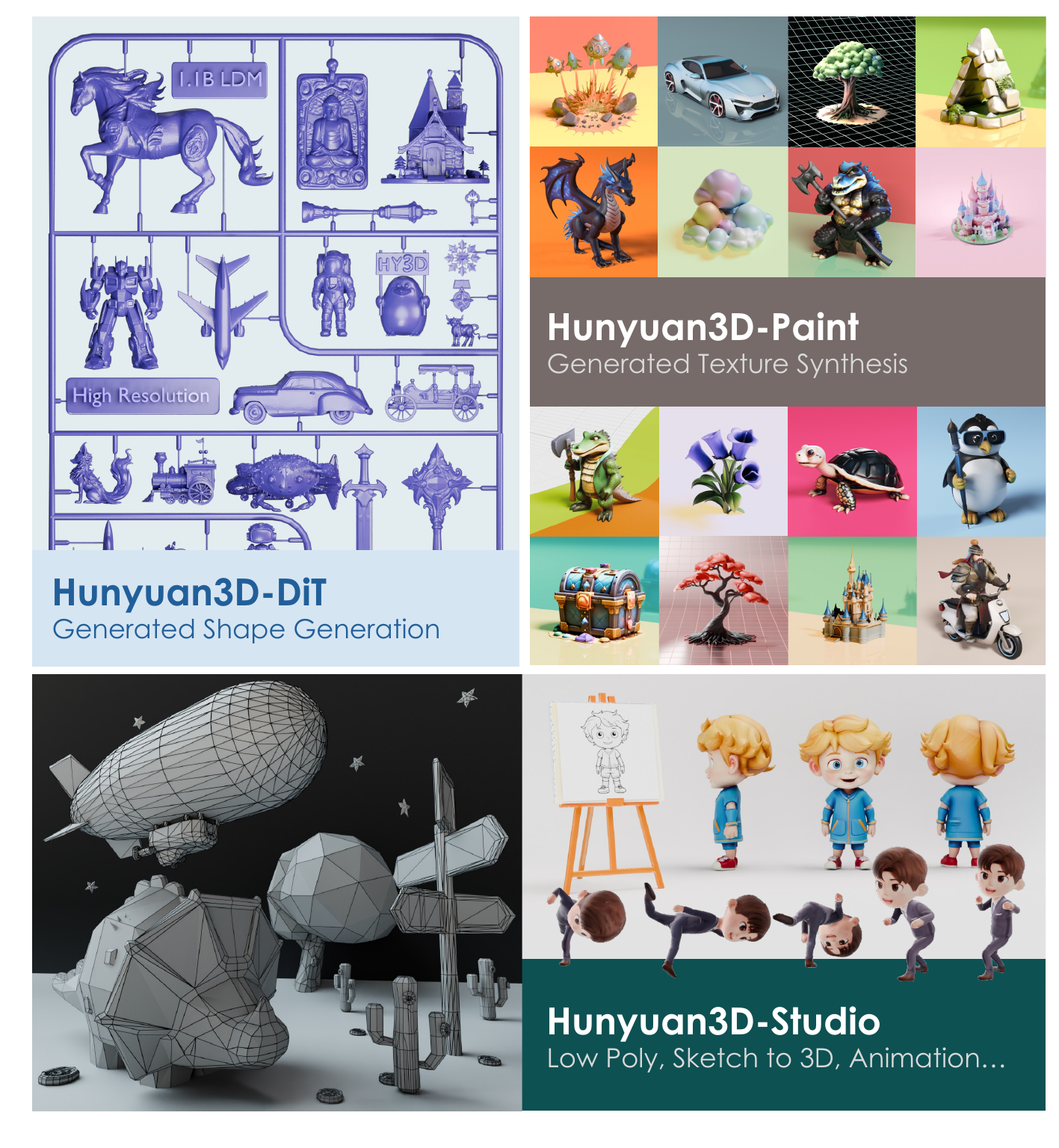}
\caption{An overall of \shortname system. }
\label{fig:sys}
\end{figure}

%% file: sections/introduction.tex
\section{Introduction}
\vspace{-2mm}
Digital 3D assets have woven themselves into the very fabric of modern life and production. In the realms of gaming and film, these assets are vibrant expressions of creators' imaginations, spreading joy and crafting immersive experiences for players and audiences alike. In the fields of physical simulation and embodied AI, 3D assets serve as essential building blocks, enabling machines and robots to mimic and comprehend the real world.
Yet, the journey of creating 3D assets is anything but straightforward; it is often a complex, time-consuming, and costly endeavor. A typical production pipeline may involve stages like sketch design, digital modeling, and 3D texture mapping, each demanding high expertise and proficiency in digital content creation software. As a result, the automated generation of high-resolution digital 3D assets has emerged as one of the most exciting and sought-after topics in recent years.

Despite the importance of automated 3D generation and rapid development in image and video generation fueled by the rise of diffusion models~\cite{ho2020denoising,rombach2022high,esser2024scaling,li2024hunyuandit,kong2024hunyuanvideo}, the field of 3D generation appears to be relatively stagnant in the era of large models and big data, with only a handful of works making gradual progress~\cite{zhang20233dshape2vecset,zhao2024michelangelo,li2024craftsman}. 
Building on the 3DShape2Vectset~\cite{zhang20233dshape2vecset}, Michelangelo~\cite{zhao2024michelangelo} and CLAY~\cite{zhang2024clay} gradually enhance shape generation performance, where CLAY is the first work to demonstrate the unprecedented potential of diffusion models in 3D asset generation.
Nevertheless, progress in the 3D domain remains limited.
As evidenced in other fields~\cite{zhang2023adding, bi2024deepseek, betker2023improving}, the prosperity of a domain in the era of large models usually relies on a strong open-source foundational model, such as Stable Diffusion~\cite{rombach2022high,podell2023sdxl,esser2024scaling} for image generation, LLaMA~\cite{touvron2023llama,touvron2023llama2,dubey2024llama} for language models, and HunyuanVideo~\cite{kong2024hunyuanvideo} for video generation. To this end, we present \emph{Hunyuan3D 2.0}, a 3D asset creation system with two strong open-sourced 3D foundation models: \emph{Hunyuan3D-DiT} for generative shape creation and \emph{Hunyuan3D-Paint} for generative texture synthesis.

\shortname features a two-stage generation pipeline, starting with the creation of a bare mesh, followed by the synthesis of a texture map for that mesh. This strategy is effective for decoupling the difficulties of shape and texture generation~\cite{hong2024lrm,yang2024hunyuan3d,lan2024ln3diff,lan2024ga} and also provides flexibility for texturing either generated or handcrafted meshes. With this architecture, our shape creation model -- Hunyuan3D-DiT, is designed as a large-scale flow-based diffusion model. As a prerequisite, we first train an autoencoder -- Hunyuan3D-ShapeVAE using advanced techniques such as mesh surface importance sampling and variational token length to capture fine-grained details on the meshes. Then, we build up a dual-single stream transformer~\cite{flux2024} on the latent space of our VAE with the flow-matching~\cite{lipman2022flow,esser2024scaling} objective. 
Our texture generation model -- Hunyuan3D-Paint is made of a novel mesh-conditioned multi-view generation pipeline and a number of sophisticated techniques for preprocessing and baking multi-view images into high-resolution texture maps. 

We performed an in-depth comparison of \shortname in relation to leading 3D generation models worldwide, including three commercial closed-source end-to-end products, an end-to-end open-sourced model Trellis~\cite{xiang2024structured}, and several separate models~\cite{chen2023text2tex,hui2024make,wu2024direct3d,zeng2024paint3d,liu2023zero,liu2024text} for shape and texture generation. We report visual and quantitative evaluation results across three dimensions: generated textured mesh, bare mesh, and texture map. We also provided user study results on 300 test cases involving 50 participants. The comparison shows the superiority of \shortname in alignment between conditional images and generated meshes, generation of fine-grained details, and human preference ratings.

%% file: sections/method.tex
\section{\shortname Architecture}

\begin{figure}
\centering
\includegraphics[width=\textwidth]{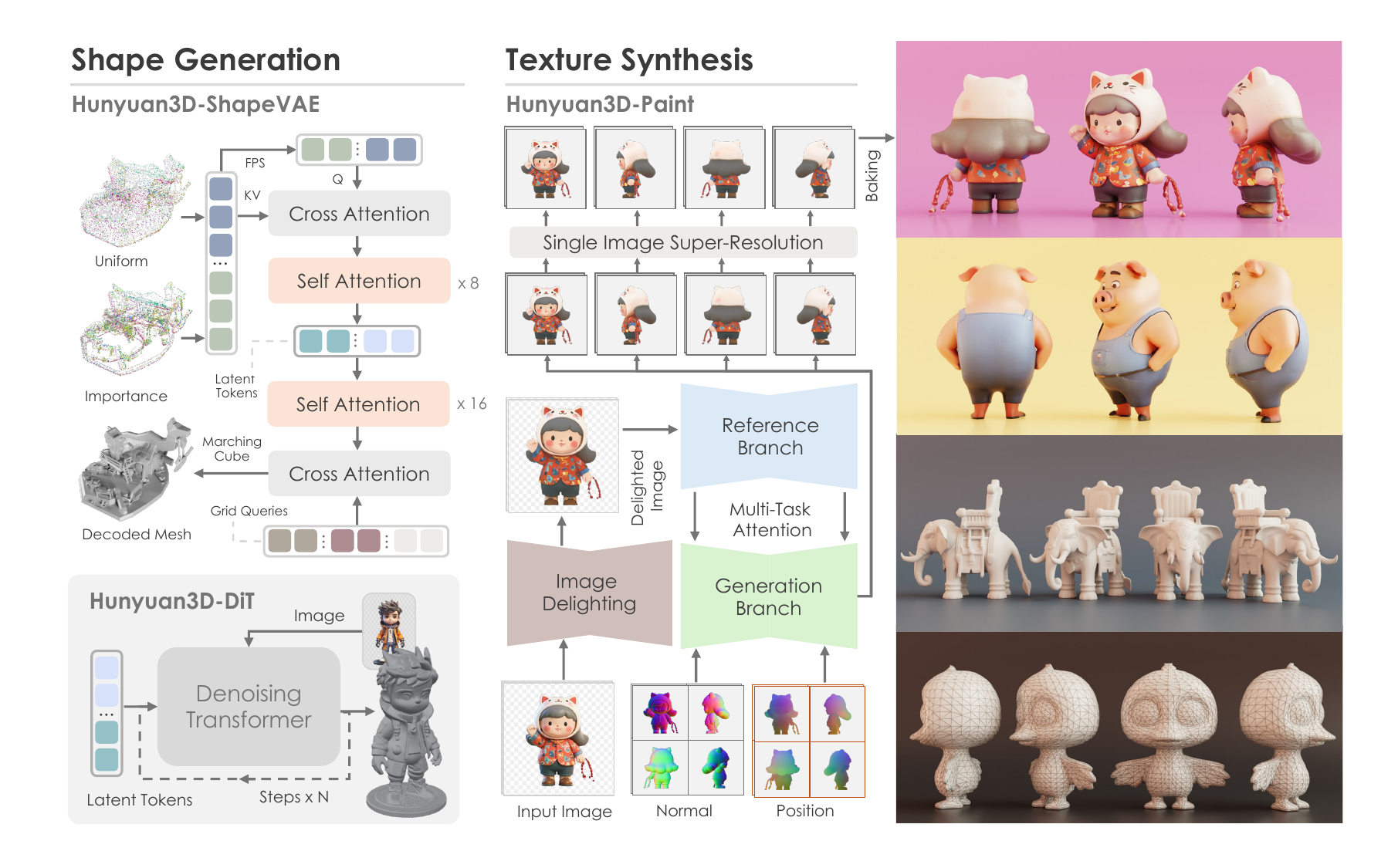}
\vspace{-7mm}
\caption{
An overall of \shortname architecture for 3D generation. It consists of two main components: Hunyuan3D-DiT for generating bare mesh from a given input image and Hunyuan3D-Paint for generating a textured map for the generated bare mesh. Hunyuan3D-Paint takes geometry conditions -- normal maps and position maps of generated mesh as inputs and generates multi-view images for texture baking.
}
\label{fig_full_pipe}
\end{figure}

In this section, we elaborate on the model architecture of \shortname, focusing on two main components: the shape generation model and the texture generation model. \cref{fig_full_pipe} illustrates the pipeline of \shortname for creating a high-resolution textured 3D asset. Given an input image, Hunyuan3D-DiT initially generates a high-fidelity bare mesh via the shape generation model. This model comprises a Hunyuan3D-ShapeVAE and a Hunyuan3D-DiT, which will be discussed in \cref{subsec:shape}. Subsequently, by leveraging strong geometric priors and the input image, we introduce Hunyuan3D-Paint as our texture generation model in \cref{subsec:texture}. This model produces self-consistent multi-view outputs, which are used for baking high-definition texture maps.

\input{sections/method/geometry}

\input{sections/method/texture}

%% file: sections/method/geometry.tex
\section{Generative 3D Shape Generation}
\label{subsec:shape}

\shortname employs the architecture of the latent diffusion model~\cite{rombach2022high,zhang20233dshape2vecset, zhao2024michelangelo,zhang2024clay} for shape generation. This design is driven by 
the successful applications of latent diffusion models in image and video generation. Specifically, our shape generation model consists of (1) an autoencoder -- Hunyuan3D-ShapeVAE (\cref{subsubsec:vae}) that compresses the shape of a 3D asset represented by polygon mesh into a sequence of continuous tokens in the latent space; (2) a flow-based diffusion model -- Hunyuan3D-DiT (\cref{subsubsec:dit}), trained on the latent space of ShapeVAE for predicting object token sequences from a user-provided image. The predicted tokens are further decoded into a polygon mesh with VAE decoder. The details of these models are illustrated below.

\subsection{Hunyuan3D-ShapeVAE}
\label{subsubsec:vae}

\begin{wrapfigure}{r}[1mm]{0pt}
\centering
\includegraphics[width=6.5cm]{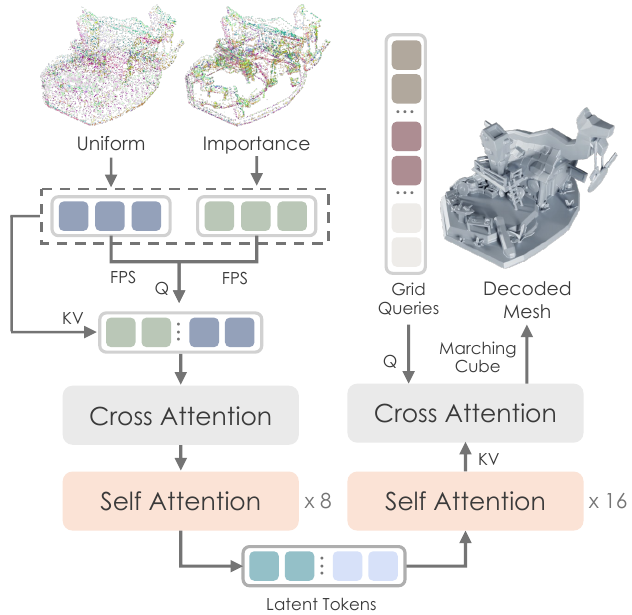}
\caption{The overall architecture of Hunyuan3D-ShapeVAE. 
Instead of only using uniform sampling on mesh surface, We have developed an importance sampling strategy to extract high-frequency detail information from the input mesh surface, such as edges and corners. This allows the model to better capture and represent the intricate details of 3D shapes.
Note that during the point query construction, the Farthest Point Sampling (FPS) operation is performed separately for the uniform point cloud and the importance sampling point cloud.}
\label{fig:vae1}
\vspace{-5mm}
\end{wrapfigure}

Hunyuan3D-ShapeVAE employs vector sets, a compact neural representation for 3D shapes proposed by 3DShape2VecSet~\cite{zhang20233dshape2vecset}, which has also been leveraged in the recent work Dora~\cite{chen2024dora}. 
Followed by Michelangelo~\cite{zhao2024michelangelo}, we use a variational encoder-decoder transformer for shape compression and decoding. Besides, we choose 3D coordinates and the normal vector of point cloud sampled from the surface of 3D shapes as inputs for the encoder and instruct the decoder to predict the Signed Distance Function (SDF) of the 3D shape, which can be further decoded into triangle mesh via the marching cube algorithm. The overall network architecture is illustrated in \cref{fig:vae1}.

\textbf{Importance Sampled Point-Query Encoder.}
The encoder $\mathcal{E}_s$ aims to extract representative features to characterize 3D shapes. To achieve this, our first design utilizes an attention-based encoder to encode point clouds uniformly sampled from the surface of a 3D shape. However, this design usually fails to reconstruct the details of complex objects. We attribute this difficulty to the variations in the complexity of regions on the shape surface. 
Therefore, in addition to uniformly sampled point clouds, we designed an importance sampling method$^*$ that samples more points on the edges and corners of the mesh, which provides more complete information for describing complex regions. 
\noindent\footnotetext{$^*$ Concurrent work~\cite{chen2024dora} also proposes similar importance sampling to improve the VAE reconstruction performance based on similar observations.}

In detail, for an input mesh, we first collect uniformly sampled surface point clouds $P_u \in \mathbb{R}^{ M \times 3 }$, and importance sampled surface point clouds $P_i \in \mathbb{R}^{ N \times 3 }$. We use a layer of cross attention, to compress the input point clouds into a set of continuous tokens via a set of point queries~\cite{zhang20233dshape2vecset}.
To obtain point queries, we apply Farthest Point Sampling (FPS) separately to $P_u$ and $P_i$ to obtain the uniform point query $Q_u \in \mathbb{R}^{ M' \times 3 }$ and the importance point query $Q_i \in \mathbb{R}^{ N' \times 3 }$. 
The final point cloud $P \in \mathbb{R}^{ (M + N) \times 3 }$ and point query $Q \in \mathbb{R}^{ (M' + N') \times 3 }$ for the cross attention are constructed by concatenating both sources.  
Then, we encode the point clouds $P$ and point queries $Q$ with Fourier positional encoding followed by a linear projection, resulting $X_p \in \mathbb{R}^{ (M+N) \times d }$ and $X_q \in \mathbb{R}^{ (M' + N') \times d }$, where $d$ is the width of the transformer.
The encoded point cloud and point query are sent to the cross attention followed by a number of self-attention layers, which helps improve the feature representation, to obtain the hidden shape representation $H_s \in \mathbb{R}^{ (M' + N') \times d }$. Since we adopt the design of variational autoencoder~\cite{kingma2013auto}, an additional linear projection is applied on $H_s$ to predict the mean $\mathrm{E}(Z_s) \in \mathbb{R}^{ (M' + N') \times d_0}$ and variance $\mathrm{Var}(Z_s) \in \mathbb{R}^{ (M' + N') \times d_0}$ of the final latent shape embedding in a token sequence, where $d_0$ is the dimension of latent shape embedding.

\textbf{Decoder.}
The decoder $\mathcal{D}_s$ reconstructs the 3D neural field from the latent shape embedding $Z_s$ from the encoder. $\mathcal{D}_s$ starts from a projection layer to transform the latent embedding from dimension $d_0$ back to the width of transformer $d$. Then, a number of self-attention layers further process the hidden embeddings, after which is another point perceiver that takes 3D grid $Q_g \in \mathbb{R}^{ (H \times W \times D) \times 3}$ as queries to obtain 3D neural field $F_g \in \mathbb{R}^{ (F_n \times W \times D) \times d}$ from the hidden embeddings. We use another linear projection on the neural field to obtain the Sign Distance Function (SDF) $F_{sdf} \in \mathbb{R}^{ (F_o \times W \times D) \times 1}$, which can be decoded into triangle mesh with marching cube algorithms. 

\textbf{Training Strategy \& Implementation.}  We employ multiple losses to supervise the model training, including (1) the reconstruction loss that computes MSE loss between predicted SDF $\mathcal{D}_s (x | Z_s)$ and ground truth $\mathrm{SDF}(x)$, and (2) the KL-divergence loss $\mathcal{L}_{KL}$ to make the latent space compact and continuous, which facilitates the training of diffusion models. Due to the dense computation required by complete SDF, the reconstruction loss is calculated as the expectation of losses on randomly sampled points in the space and shape surface. The overall training loss $\mathcal{L}_r$ can be written as,
\begin{equation}
    \mathcal{L}_r = \mathbb{E}_{ x \in \mathbb{R}^3 } [ \mathrm{MSE} ( \mathcal{D}_s (x | Z_s), \mathrm{SDF}(x) ) ] + \gamma \mathcal{L}_{KL}
\end{equation}
where $\gamma$ is the loss weight of KL loss.
During training, we also utilize a multi-resolution strategy to speed up model convergence, where the length of the latent token sequence is randomly sampled from a predefined set. A shorter sequence reduces the computation cost, and a longer sequence facilitates the reconstruction quality. The longest sequence length is 3072 in our released version, which could support high-resolution shape generation with fine-grained and sharp details.

\subsection{Hunyuan3D-DiT}
\label{subsubsec:dit}
Hunyuan3D-DiT is a flow-based diffusion model aimed at producing high-fidelity and high-resolution 3D shapes according to given image prompts.

\begin{figure}[t]
\centering
\includegraphics[width=\textwidth]{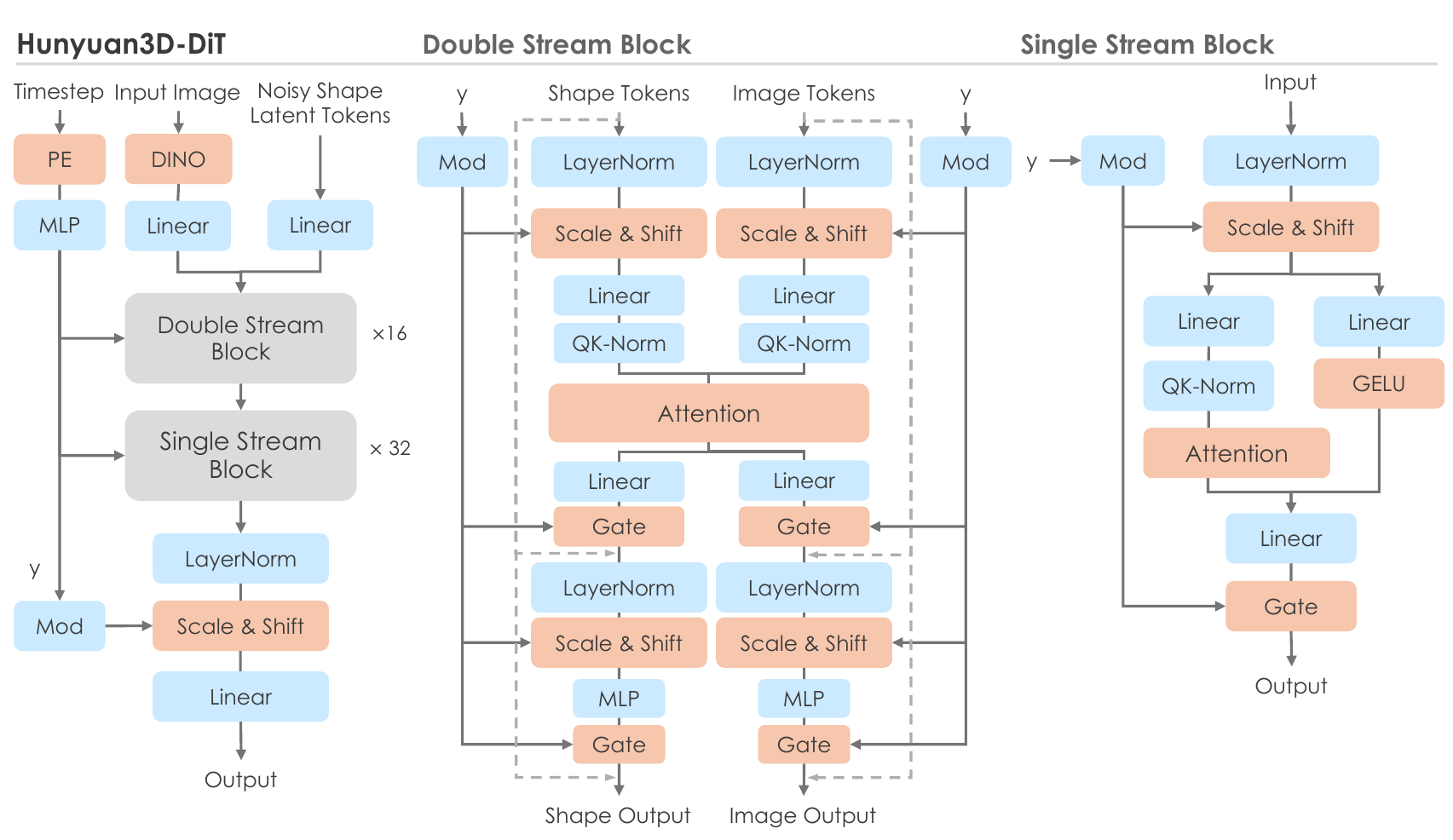}
\vspace{-7mm}
\caption{Overview of Hunyuan3D-DiT. It adopts a transformer architecture with both double- and single-stream blocks. This design benefits the interaction between modalities of shape and image, helping our model to generate bare meshes with exceptional quality. (Note that the orange blocks have no learnable parameters, the blue blocks contain trainable parameters, and the gray blocks indicate a module composed of more details.)}
\label{fig:dit}
\end{figure}

\textbf{Network Structure.} Inspired by FLUX~\cite{flux2024}, we adopt a dual- and single-stream network structure as in \cref{fig:dit}. 
In dual-stream blocks, latent tokens and condition tokens are processed with separate QKV projections, MLP, and \etc, but interact within an attention operation. 
In single-stream blocks, latent tokens and condition tokens are concatenated and processed by spatial attention and channel attention in parallel.
We only use the embedding of the timestep for the modulation modules. Besides, we omit the positional embedding of the latent sequence as the specific latent token of our ShapeVAE in the sequence does not correspond to a fixed location in the 3D grid. Instead, the content of our 3D latent tokens themselves is responsible for figuring out the position/occupancy of the generated shape in the 3D grid, which is different from image/video generation where their tokens are responsible for predicting content in a specific location in 2D/spatial-temporal grid. 

\textbf{Condition Injection.} We employ a pre-trained image encoder to extract conditional image tokens of the patch sequence including the head token at the last layer. To capture the fine-grained details in the image, we utilize a large image encoder -- DINOv2 Gaint~\cite{oquab2023dinov2} and large input image size -- $518 \times 518$. Besides, we also remove the background of the input image, resize the object to a unified size, reposition the object to the center, and fill the background with white, which helps to remove the negative impact of the background and increase the effective resolution of the input image.

\textbf{Training \& Inference.} We utilize flow matching objective~\cite{lipman2022flow,esser2024scaling} for training our model. Specifically, flow matching first defines a probability density path between Gaussian distribution and data distribution, then, the model is trained to predict the velocity field $u_t = \frac{x_t}{d_t}$ that drifting sample $x_t$ towards data $x_1$. In our case, we adopt the affine path with the conditional optimal transport schedule specified in ~\cite{lipman2024flowmatchingguidecode}, where $x_t = ( 1-t ) \times x_0 + t \times x_1$, $u_t = x_1 - x_0$. Therefore, the training loss is formulated as,
\begin{equation}
    \mathcal{L} = \mathbb{E}_{t,x_0,x_1} [ \parallel u_\theta(x_t,c,t) - u_t \parallel_2^2 ],
\end{equation}
where $t \sim \mathbb{U}(0,1)$ and $c$ denotes model condition. 
During the inference phase, we first randomly sample a start point $x_0 \sim \mathbb{N}(0,1)$ and employ a first-order Euler ordinary differential equation (ODE) solver to solve $x_1$ with our diffusion model $u_\theta(x_t,c,t)$.

%% file: sections/method/texture.tex
\section{Generative Texture Map Synthesis}
\label{subsec:texture}

Given a 3D mesh without texture and an image prompt, we aim to generate a high-resolution and seamless texture map. 
The texture map should closely conform to the image prompt in the visible region, exhibit multi-view consistency, and maintain harmonious with the input mesh.

\begin{figure*}
    \centering
    \includegraphics[width=\linewidth]{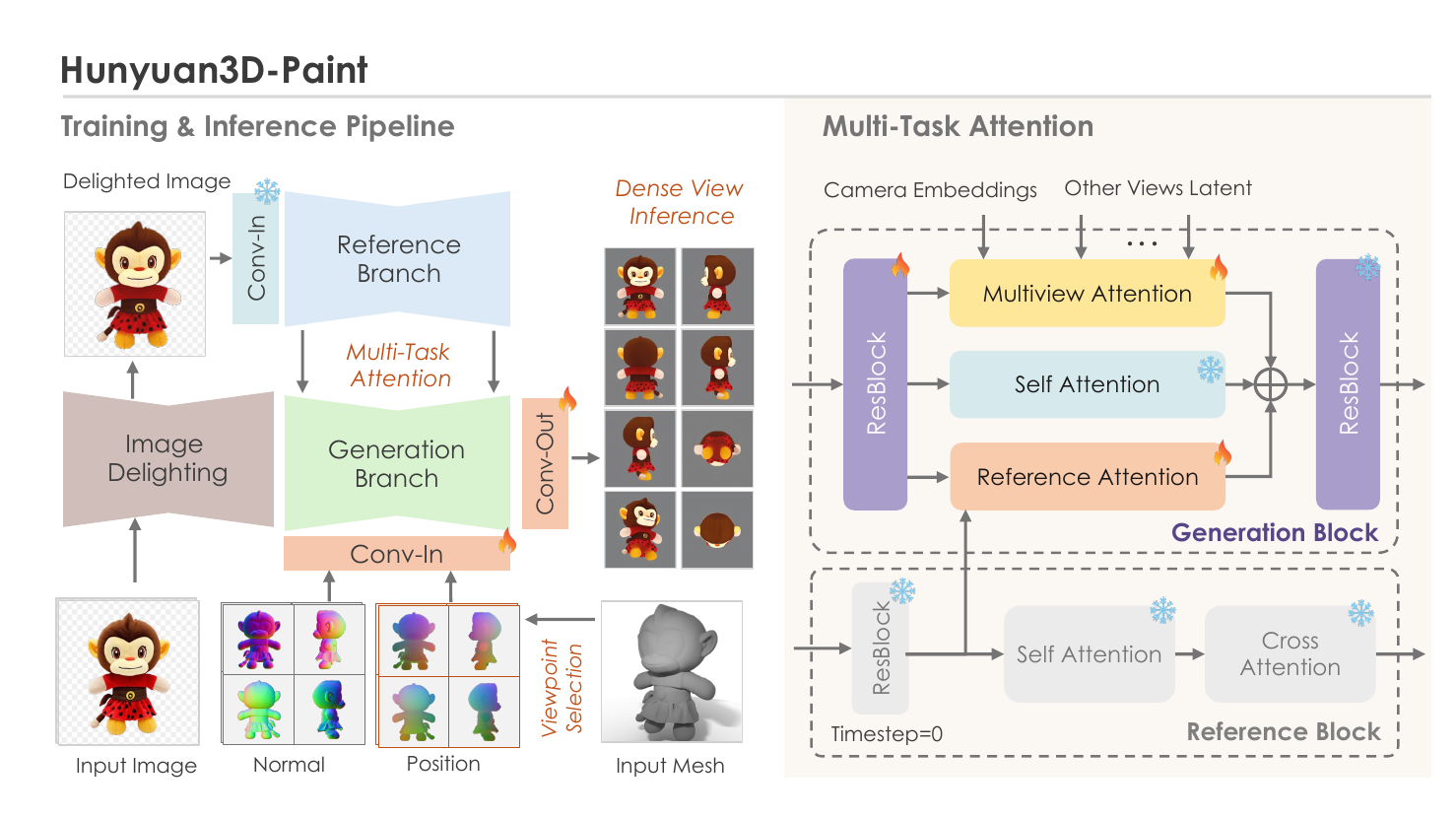}
    \vspace{-5mm}
    \caption{Overview of Hunyuan3D-Paint. We leverage an image delighting module to convert the input image to an unlit state to produce light-invariant texture maps.
The system features a double-stream image conditioning reference-net, which provides faithfully conditional image features to the model. Furthermore, it facilitates the production of texture maps that conform closely to the input image.
The multi-task attention module ensures that the model synthesizes multi-view consistent images. This module maintains the coherence of all generated images while adhering to the input.}
    \label{fig:texture_overview}
\end{figure*}

To achieve these objectives, we employ a three-stage framework, including a pre-processing stage (\cref{subsubsec:preprocess}), a multi-view image synthesis stage (\cref{subsubsec:MVD}, Hunyuan3D-Paint), and a texture baking stage based on dense multi-view inference (\cref{subsubsec:dense_view_inference}). 
Other details related to model training and text- and image-to-texture pipeline are provided in \cref{subsubsec:implementation}. \cref{fig:texture_overview} illustrates the complete pipeline of our texture map synthesis method.

\subsection{Pre-processing}
\label{subsubsec:preprocess}

\noindent\textbf{Image Delighting Module.}
The reference image typically exhibits pronounced and varied illumination and shadow, whether collected by the user or generated by T2I models. 
Directly inputting such images into the multi-view generation framework can cause illumination and shadows to be baked into the texture maps. 
To address this issue, we leverage a delighting procedure on the input image via an image-to-image approach~\cite{brooks2022instructpix2pix} before multi-view generation. Specifically, to train such an image delighting model, we collect a large-scale 3D dataset and render it under the illumination of a random HDRI environmental map and an even white light to form the corresponding pair-wise image data. Benefiting from this image delighting model, our multi-view generation model can be fully trained on white-light illuminated images, enabling an illumination-invariant texture synthesis.

\noindent\textbf{View Selection Strategy.}
In practical applications, to reduce the costs of texture generation (i.e., generate the largest area of texture with the minimum number of viewpoints), we employ a geometry-aware viewpoint selection strategy to support effective texture synthesis. 
By considering the coverage of the geometric surface, we heuristically select 8 to 12 viewpoints for inference. 
Initially, we fix 4 orthogonal viewpoints as basis since they cover most parts of the geometry. Subsequently, we iteratively add novel viewpoints using a greedy search approach.
The specific process is illustrated in Algorithm~\ref{alg:view-selection}, and the coverage function in the algorithm is defined as:
\begin{equation}
    \mathcal{F}(v_i, \mathbb{V}_{s}, \mathbf{M}) = \mathcal{A}_{area}\left\{\mathcal{UV}_{\text{cover}}(v_i, \mathbf{M}) \setminus \left[ \mathcal{UV}_{\text{cover}}(v_i, \mathbf{M}) \cap \left(\bigcup_{s \in \mathbb{V}_{s}} \mathcal{UV}_{\text{cover}}(v_s, \mathbf{M}) \right)\right]\right\}
\end{equation}
where $\mathcal{UV}_{\text{cover}}(v, \mathbf{M})$ is a function that returns the set of covering texels in UV space based on the input view $v$ and mesh geometry $\mathbf{M}$, and $\mathcal{A}_{area}(\cdots)$ is a function that calculates the coverage area according to the given set of covering texels. 
This approach encourages the multi-view generation model to focus on viewpoints with more unseen regions, together with the dense-view inference, alleviating the burden of post-processing (i.e., texture inpainting).

\begin{algorithm}
\caption{View Selection Algorithm}
\label{alg:view-selection}
\begin{algorithmic}[1]
\State \textbf{Input:}
\State \quad Initialize a selected viewpoint set $\mathbb{V}_{s}$, a reference viewpoint set $\mathbb{V}_{r}$, a coverage function $\mathcal{F}$, and a mesh $\mathbf{M}$.
\State \quad Set $N_{max} = 12$ as the maximum number of iterative searches, and $N_{fixed} = 4$ as the initial fixed number of viewpoints.
\For{$i = N_{fixed}$ \textbf{to} $N_{max} - 1$}
    \State \textbf{Set} $c_{max} = -1$
    \State \textbf{Set} $v_{max} = -1$
    \For{$v_i \in \mathbb{V}_{r}$}
        \State \textbf{Set} $c_{cur} = \mathcal{F}(v_i, \mathbb{V}_{s}, \mathbf{M})$
        \If{$c_{cur} > c_{max}$}
            \State $c_{max} = c_{cur}$
            \State $v_{max} = v_i$ 
        \EndIf
    \EndFor
    \State $\mathbb{V}_{r} \gets \mathbb{V}_{r} \setminus \{ v_{max} \}$ 
    \State $\mathbb{V}_{s} \gets \mathbb{V}_{s} \cup \{ v_{max} \}$ 
\EndFor
\end{algorithmic}
\end{algorithm}
\vspace{-2mm}

\subsection{Hunyuan3D-Paint}
\label{subsubsec:MVD}
Geometry-conditioned multi-view image generation is a key component in the texture synthesis framework. In the context of image-guided texture synthesis, the design of the multi-view image generation must be meticulously crafted to achieve image alignment, geometric following, and multi-view consistency.
These functionalities are realized in Hunyuan3D-Paint according to a bunch of techniques containing a double-stream image conditioning reference-net, a multi-task attention mechanism, and a strategy for geometry and view conditioning.

\noindent\textbf{Double-stream Image Conditioning Reference-Net.}
In the aim of the image following, Hunyuan3D-Paint implements a reference-net conditioning approach~\cite{RefNet2023}, upon which we develop a series of alternative solutions.
Specifically, rather than a noisy feature synchronized with the generation branch, we directly feed the original VAE feature of the reference image into the reference branch to maintain image details as much as possible. 
Since the feature is noiseless, we set the timestep of the reference branch to 0 to maintain the input image information faithfully.
On the other hand, to regularize potential style bias introduced by the 3D rendering dataset, we abandon the shared-weight reference-net as in~\cite{tang2025mvdiffusion++, shi2023zero123++} and freeze the weights of the original SD2.1 weights (which serves as the base model for our multi-view generation). We have found that the fixed-weights reference-net serves as a soft regularization that anchors the generated image distribution, preventing it from drifting away towards the rendered image distribution, significantly improving the performance on real-world image conditioning. Together, these two schemes form the double-stream image conditioning strategy. We leverage the zero-noised double-stream image conditioning reference-net by capturing a feature cache prior to each self-attention module. This cache is then fed into the multi-view diffusion model via a reference attention module.

\noindent\textbf{Multi-task Attention Mechanism.}
We introduce two additional attention modules alongside the original self-attention to enable the image diffusion model to generate multi-view images guided by a reference image. 
The reference attention module integrates the reference image into the multi-view diffusion process. In contrast, the multi-view attention module ensures consistency across the generated views.
To mitigate potential conflicts arising from these multi-functionalities, we design the additional two distinct attention modules in a parallel structure (illustrated in \cref{fig:texture_overview}), which can be expressed as:
\begin{equation}
Z_{MVA} = Z_{SA} + \lambda_{ref} \cdot \text{Softmax}\left(\frac{Q_{ref}K_{ref}^T}{\sqrt{d}}\right) V_{ref} + \lambda_{mv} \cdot \text{Softmax}\left(\frac{Q_{mv}K_{mv}^T}{\sqrt{d}}\right) V_{mv}
\end{equation}
where $Z_{SA}$ represents the feature calculated by the original frozen-weight self-attention, and $Q_{ref}, K_{ref}, V_{ref}$ and $Q_{mv}, K_{mv}, V_{mv}$ are the Query, Key, and Value projected features of reference attention and multi-view attention, respectively.

\noindent\textbf{Geometry and View Conditioning.}
\label{subsubsubsec:view_conditioning}
Following geometry is another unique feature in texture map synthesis. 
To enable effective training, we opt for an easy implementation of directly concatenating the geometry conditions with noise. 
Specifically, we first input the multi-view canonical normal maps and canonical coordinate maps (CCM)—two view-invariant geometry conditions we utilize—into a pre-trained Variational Autoencoder (VAE) to obtain geometric features. These features are then concatenated with latent noise and fed into the channel-extended input convolution layer of the diffusion model.

Other than geometry conditioning, we adopt a learnable camera embedding in our pipeline to boost the viewpoint clue for the multi-view diffusion model.
Specifically, we assign a unique unsigned integer to each pre-defined viewpoint and set up a learnable view embedding layer to map the integer to a feature vector, which is then injected into the multi-view Diffusion model.  
We have found in our experiments that combining the geometry conditioning with a learnable camera embedding yields the best performance.

\subsection{Texture Baking}
\label{subsubsec:dense_view_inference}
\noindent\textbf{Dense-view inference.}
Potential self-occlusion is a significant challenge in the context of texture synthesis within a multi-view image generation framework, particularly when handling irregular geometries produced by shape generative models. This issue necessitates modeling texture synthesis as a two-stage framework. The first stage focuses on multi-view image generation, while the second stage involves non-trivial inpainting to fill the holes caused by self-occlusion. In our texture-map synthesis framework, we alleviate the burden on the second stage of inpainting by facilitating dense-view inference during the multi-view generation stage. To enable effective dense-view inference, a view dropout strategy is introduced as a flexible training mechanism that allows the model to encounter all of the pre-set viewpoints, thereby enhancing its 3D perception capabilities and generalization. Specifically, we randomly select 6 viewpoints from a total of 44 pre-set viewpoints to serve as a batch input to the multi-view diffusion backbone network. During the inference phase, our framework is able to output the images of any specified viewpoints, supporting dense-view inference.

\noindent\textbf{Single Image Super-resolution.}
To enhance texture quality, we apply a pre-trained single-image super-resolution model~\cite{wang2018esrgan} to each generated image from different viewpoints. Experiments have demonstrated that this single-image super-resolution approach maintains consistency among multi-views, as it does not introduce significant variations to the images.

\noindent\textbf{Texture Inpainting.}
After unwrapping the synthesized dense multi-view images into a texture map, a small set of patches in the UV texture remain that are not fully covered. To address this issue, we employ an intuitive inpainting approach. 
First, we project the existing UV texture into vertex texture. 
Then, we query each UV texel's texture by computing a weighted sum of the textures from the connected, textured vertices.
The weights are set to be the reciprocal of the geometric distances between the texels and the vertices.

\subsection{Implementation Details}
\label{subsubsec:implementation}
\noindent\textbf{Model Training.}
For training the multi-view image generation framework, we start by inheriting the ZSNR checkpoint of the Stable Diffusion 2 v-model~\cite{lin2024common}. We train our multi-view diffusion model using a self-collected large-scale 3D dataset. 
Multi-view images are rendered under the illumination of an even white light to accommodate our delighting model. 
Specifically, we render the reference image with a random azimuth and a fixed range of elevation from -20 to 20 degrees. This variation disrupts the consistency between the reference and generated images, thereby increasing the robustness of our texture generation framework.
We directly train on 512 × 512 resolution with a total of 80,000 steps, a batch size of 48, and a learning rate of $5 \times 10^{-5}$. We use 1000 warm-up steps and the "trailing" scheduler proposed by ZSNR.

\noindent\textbf{General Text- and Image-to-Texture.}
It is worth noting that Hunyuan3D-Paint not only generates high-quality texture maps for generated meshes but also supports arbitrary texture generation guided by any text or image input provided by the user for any geometric model. 
To achieve this, we leverage advanced T2I models and corresponding conditional generation modules, such as ControlNet~\cite{zhang2023adding} and IP-Adapter~\cite{ye2023ip-adapter}, to generate input images that align with geometric shapes based on user-provided text or image prompts. 
Benefitting from this paradigm, we are capable of texturing any specified geometry with arbitrary images, whether they are matched or mismatched.
An application of using different images to texture the same geometry, dubbed as re-skinning, is illustrated in \cref{fig:skinning}.

%% file: sections/experiments.tex
\section{Evaluations}

To thoroughly evaluate the performance of \shortname, we conducted experiments from three perspectives: (1) 3D Shape Generation (including Shape Reconstruction and Shape Generation), (2) Texture map synthesis, and (3) Textured 3D assets generation.

\input{sections/experiments/geometry}

\input{sections/experiments/texture}

\input{sections/experiments/textured-assets}

\input{sections/experiments/applications}

%% file: sections/experiments/geometry.tex
\subsection{3D Shape Generation}

Shape generation is crucial for 3D generation, as high-fidelity and high-resolution bare meshes form the foundation for downstream tasks. 
In this section, we compare and evaluate the capability of 3D shape generation in \shortname from two perspectives: shape reconstruction and shape generation.

\textbf{Baselines.}
We compare the reconstruction performance of Hunyuan3D-ShapeVAE with 3DShape2VecSet~\cite{zhang20233dshape2vecset}, Michelangelo~\cite{zhao2024michelangelo}, and Direct3D~\cite{wu2024direct3d}. 
The mentioned methods represent the state-of-the-art ShapeVAE architecture, and the core differences are neural representations, where 3DShape2VecSet uses a downsampled vector set, point query; Michelangelo utilizes a learnable vector set, learnable query; Direct3D leverages learnable triplane; Hunyuan3D-ShapeVAE employees point query with importance sampling. Note that, except Direct3D, which requires a 3072 token length (suffering significant performance degeneration when reducing token length), all VAE models compare by 1024 token length.
Hunyuan3D-DiT is compared with several state-of-the-art baselines. Open-source baselines are Michelangelo~\cite{zhao2024michelangelo}, Craftsman 1.5~\cite{li2024craftsman}, and Trellis~\cite{xiang2024structured}. Closed-source baselines are Shape Model 1, Shape Model 2, and Shape Model 3.

\textbf{Metrics.}
We employ the Intersection of Union (IoU) to measure the reconstruction performance. Specifically, we compute randomly sampled volume points IoU (V-IoU) and near-surface region IoU (S-IoU) to reflect the reconstruction performance comprehensively. 
To evaluate shape generative performance, we employ ULIP~\cite{xue2023ulip} and Uni3D~\cite{zhou2023uni3d} to compute the similarity between the generated mesh and input images (ULIP-I and Uni3D-I) and the similarity between the generated mesh and images prompt synthesizing by the vision language model~\cite{chen2024internvl} (ULIP-T and Uni3D-T).

\textbf{Shape Reconstruction Comparisons.}
\begin{figure*}
    \centering
    \includegraphics[width=\textwidth]{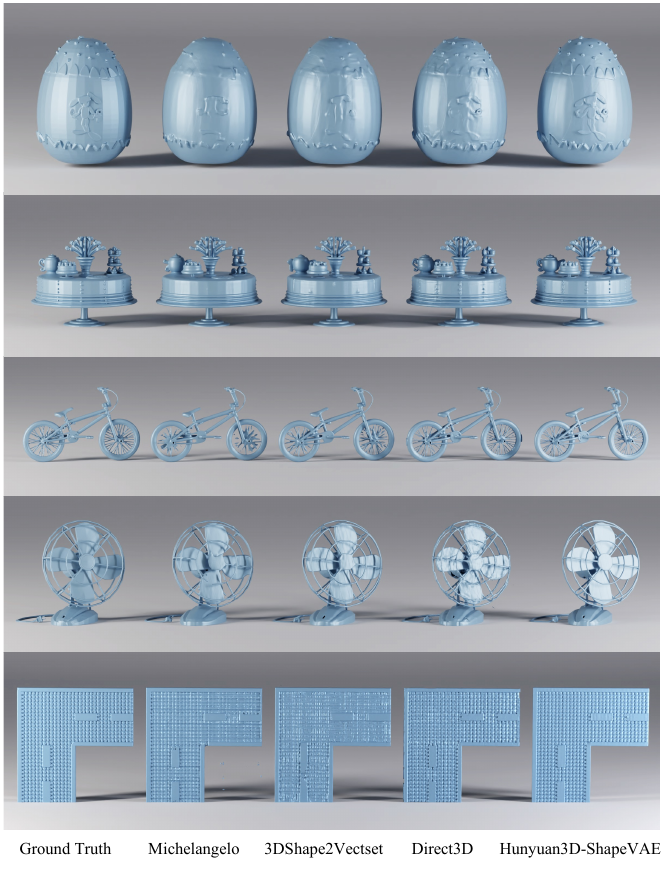}
\caption{Visual comparisons.
We illustrate the reconstructed mesh (blue paint aims to show more details) in the figure, which showcases that only Hunyuan3D-ShapeVAE reconstructs mesh with fine-grained surface details and neat space. (Better viewed by zooming in.)}
    \label{fig:recon}
\end{figure*}
The Numerical comparison of shape reconstruction is shown in \cref{tab:recon}. According to the table, Hunyuan3D-ShapeVAE overwhelms all baselines. 
Comparisons among Hunyuan3D-ShapeVAE, 3DShape2VecSet, and Michelangelo demonstrate the effectiveness of the importance sampling strategies.
\cref{fig:recon} illustrates the visual comparison of shape reconstruction, which 
shows that Hunyuan3D-ShapeVAE could faithfully recover the shape with fine-grained details and produce neat space without any floaters.
\input{tables/shape/reconstruction}

\begin{figure*}
    \centering
    \includegraphics[width=\linewidth]{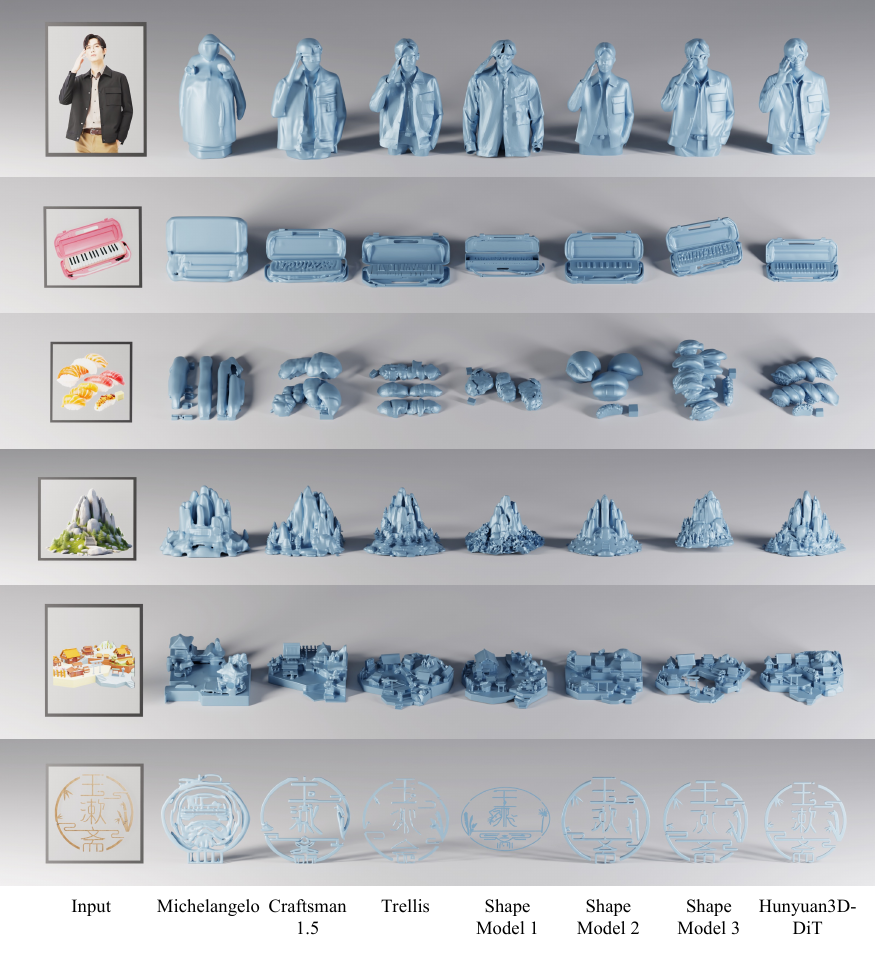}
    \vspace{-2mm}
    \caption{Visual comparisons.
We display the input image and the generated bare mesh (blue paint aims to show more details) from all methods in the figure. The human faces and piano keys show that Hunyuan3D-DiT could synthesize detailed surface bumps, maintaining completeness. Several scenes or logos demonstrate that Hunyuan3D-DiT could generate intricate details. (Better viewed by zooming in.)}
    \label{fig:shapegen}
\end{figure*}

\textbf{Shape Generation Comparisons.} 
\cref{tab:shapegen_transposed} shows the numerical comparison between Hunyuan3D-DiT and competing methods, which indicates that Hunyuan3D-DiT produces the most condition following results.
Furthermore, according to the visual comparison in \cref{fig:shapegen}, results from Hunyuan3D-DiT follow the image prompt most, including clear human faces, surface bumps, logo texts, and layouts.
Meanwhile, the generated bare mesh is holeless, which supports a solid basis for downstream tasks.
\input{tables/shape/shape-gen}

%% file: tables/shape/reconstruction.tex
\begin{table}
\resizebox{1\columnwidth}{!}{%
\centering
\begin{tabular}{rcccccccc}
\hline
            & 3DShape2VecSet~\cite{zhang20233dshape2vecset} & Michelangelo~\cite{zhao2024michelangelo} & Direct3D~\cite{wu2024direct3d} & Hunyuan3D-ShapeVAE (Ours)    \\ \hline
V-IoU($\uparrow$)       & 87.88\%    & 84.93\%     & \underline{88.43}\%    & \textbf{93.6\%}       \\ 
S-IoU($\uparrow$)       & 80.66\%    & 76.27\%     & \underline{81.55}\%    & \textbf{89.16\%}       \\        \hline
\end{tabular}
}
\caption{Numerical comparisons. 
We evaluate the reconstruction performance of Hunyuan3D-ShapeVAE and baselines based on volume IoU (V-IoU) and Surface (S-IoU). 
The results indicate Hunyuan3D-ShapeVAE overwhelms all baselines in the reconstruction performance.}
\label{tab:recon}
\end{table}

%% file: tables/shape/shape-gen.tex

\begin{table}
\centering
\small
\begin{tabular}{rccccccc}
\hline
                        & ULIP-T($\uparrow$) & ULIP-I($\uparrow$) & Uni3D-T($\uparrow$)& Uni3D-I($\uparrow$) \\ \hline
Michelangelo~\cite{zhao2024michelangelo} & 0.0752 & 0.1152 & 0.2133 & 0.2611 \\
Craftsman 1.5~\cite{li2024craftsman}    & 0.0745 & 0.1296 & 0.2375 & 0.2987 \\
Trellis~\cite{xiang2024structured}      & 0.0769 & 0.1267 & 0.2496 & 0.3116 \\
Shape Model 1 & \underline{0.0799} & 0.1181 & 0.2469 & 0.3064 \\
Shape Model 2                & 0.0741 & \textbf{0.1308} & 0.2464 & 0.3106 \\
Shape Model 3                 & 0.0746 & 0.1284  & \underline{0.2516} & \underline{0.3131} \\
Hunyuan3D-DiT (Ours)                    & \textbf{0.0771} & \underline{0.1303} & \textbf{0.2519} & \textbf{0.3151} \\ \hline
\end{tabular}
\vspace{3mm}
\caption{Numerical comparisons.
By evaluating the shape generation performance on ULIP-T/I, Uni3D-T/I, demostrating Hunyuan3D-DiT could produce the most condition followed results.}
\label{tab:shapegen_transposed}
\end{table}

%% file: sections/experiments/texture.tex
\subsection{Texture Map Synthesis}
\label{exp:tex}

As texture maps directly influence the visual appeal of textured 3D assets, we conduct comprehensive text-conditioned texture map synthesis experiments to validate the performance of Hunyuan3D-Paint.

\textbf{Baselines.}
We compare Hunyuan3D-Paint with the following texture generation methods, including TEXTure~\cite{richardson2023texture}, Text2Tex~\cite{chen2023text2tex}, SyncMVD~\cite{liu2024text}, Paint3D~\cite{zeng2024paint3d}, and TexPainter~\cite{zhang2024texpainter}.
All the baselines leverage geometric and diffusion priors to facilitate the overall generation quality of texture maps.

\textbf{Metrics.}
We apply several frequently used image-level metrics to enable a fair comparison of texture map generation. 
Specifically, we leverage a CLIP-version of Fréchet Inception Distance $FID_{CLIP}$ to compute the distance between the rendering of the generated textured map in semantic perspectives.  We use the implementation of Clean-FID~\cite{parmar2021cleanfid}.
Besides, the recently introduced CLIP Maximum-Mean Discrepancy (CMMD)~\cite{jayasumana2024rethinking} is utilized to serve as another important criterion, which is a more accurate measurement of images with rich details. 
In addition to these two metrics, we also use CLIP-score~\cite{radford2021learning} to validate semantic alignment between renderings of the generated texture map and given prompt and LPIPS~\cite{zhang2018perceptual} to estimate the consistency between renderings of the generated texture map and ground-truth images.

\begin{figure*}
    \centering
    \includegraphics[width=\columnwidth]{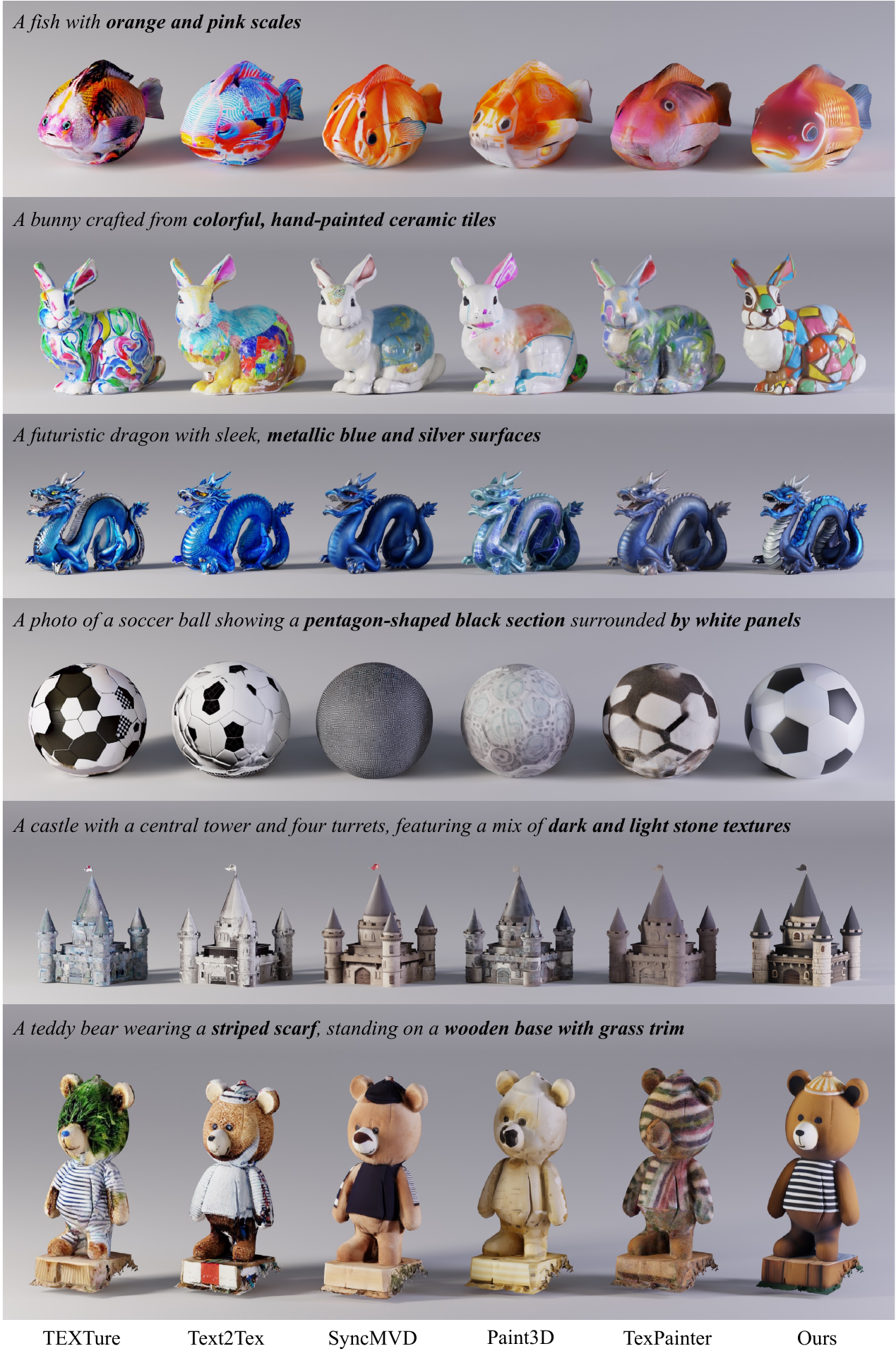}
\vspace{-2mm}
\caption{Visual comparisons.
We demonstrate several generated texture maps on different bare meshes. The fish and rabbit texture map showcases that Hunyuan3D-Paint produces the most text-conforming results.
The football indicates that our model could synthesize seamless and clean texture maps.
Moreover, Hunyuan3D-Paint could generate complex texture maps, like the castle and bear.
(Better viewed by zooming in.)}
    \label{fig:texgen}
\end{figure*}

\textbf{Comparisons.}
The numerical comparison of text-to-texturing is shown in \cref{tab:text2texture}, showcasing that Hunyuan3D-Paint achieves the best generative quality and semantic following.
The visual comparison refers to \cref{fig:texgen}.
The fish and rabbit show that our model produces the most condition-following results. And the football demonstrates the ability of Hunyuan3D-Paint to produce clear texture maps. The texture map of the castle and bear contains rich texture patterns, showcasing that our model can produce intricate details.

\begin{figure*}
    \centering
    \includegraphics[width=\columnwidth]{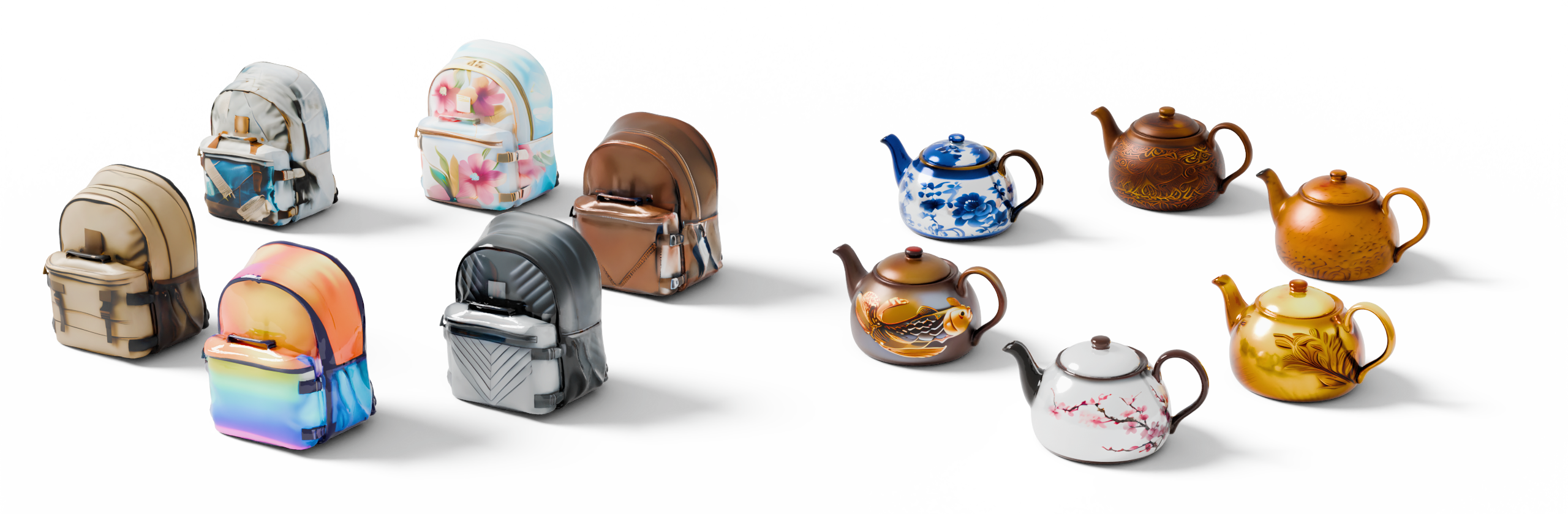}
    \vspace{-5mm}
\caption{Visual results.
We generate different texture maps for two meshes, and the results validate the performance of Hunyuan3D-Paint on texture reskinning. (Better viewed by zooming in.)}
    \label{fig:skinning}
\end{figure*}

\textbf{Applications.}
All generated texture maps are seamless and lighting-invariant.
Moreover, Hunyuan3D-Paint is flexible to produce various texture maps for bare mesh or hand-crafted mesh according to different prompts. As shown in the \cref{fig:skinning}, our model produces different texture maps for a mesh with seamless and intricate details.

\input{tables/texture/text-to-texture}

%% file: tables/texture/text-to-texture.tex
\begin{table}
\centering
\small
\begin{tabular}{rcccccccc}
\hline
        & CMMD($\downarrow$) & FID$_{CLIP}$($\downarrow$) & CLIP-score($\uparrow$) & LPIPS($\downarrow$)    \\ \hline
TEXTure~\cite{richardson2023texture}  & 3.047   & 35.75     & 0.8499    & 0.0076       \\ 
Text2Tex~\cite{chen2023text2tex}  & 2.811    & 31.72     & 0.8680    & 0.0071       \\
SyncMVD~\cite{liu2024text}  & 2.584    & 29.93     & 0.8751    & 0.0063      \\
Paint3D~\cite{zeng2024paint3d}  & 2.810    & 30.29     & 0.8724    & 0.0063        \\
TexPainter~\cite{zhang2024texpainter}  & \underline{2.483}    & \underline{28.83}     & \underline{0.8789}    & \underline{0.0062}       \\
Hunyuan3D-Paint (Ours)  & \textbf{2.318}    & \textbf{26.44}     & \textbf{0.8893}    & \textbf{0.0059}        \\\hline
\end{tabular}
\vspace{3mm}
\caption{Numerical comparisons.
We compare Hunyuan3D-Paint with baselines on various metrics, and the results indicate that our model could produce the most condition-conforming texture maps.}
\label{tab:text2texture}
\end{table}

%% file: sections/experiments/textured-assets.tex
\subsection{Textured 3D Assets Generation}

In this section, we evaluate the generated textured 3D assets for reflecting the end-to-end generation capabilities of \shortname.

\textbf{Baselines.}
We compare \shortname against leading models in the field, including open-source model Trellis~\cite{xiang2024structured} and closed-source models Model 1, Model 2, and Model 3.

\textbf{Metrics.}
We mainly measure the generative quality of textured 3D assets by their renderings. 
Similar to \cref{exp:tex}, we employ $FID_{CLIP}$ to compute the image content distance, CLIP-score to reflect semantic alignment, CMMD to measure the similarity in the image details, and LPIPS to evaluate the consistency between rendering from generated textured 3D assets and given image prompts.

\begin{figure*}[t]
    \centering
    \includegraphics[width=\linewidth]{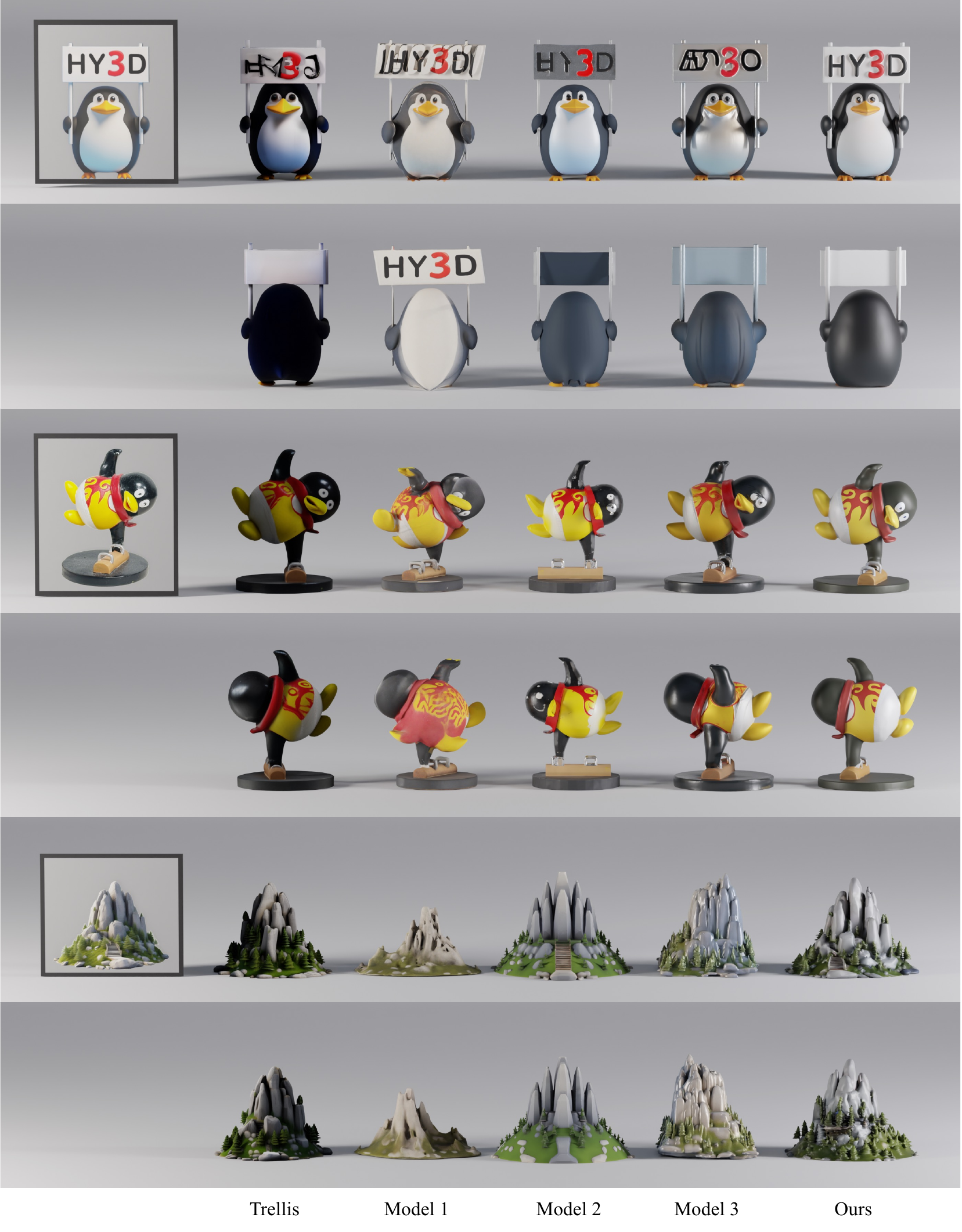}
\caption{Visual comparisons.
The first case reflects that \shortname could synthesize detailed surface bumps and correct texture maps.
The second penguin showcases our model's ability to handle complex actions.
The last mountain demonstrates that Hunyuan3D-DiT could produce intricate structures, and Hunyuan3D-Paint can synthesize vivid texture maps.
(Better viewed by zooming in.)}
    \label{fig:e2e}
\end{figure*}

\textbf{Comparisons.}
The numerical results reported in the \cref{tab:e2e} indicate that \shortname surpasses all baselines in the quality of generated textured 3D assets and the condition following ability.
The illustration in the \cref{fig:e2e} demonstrates that \shortname produces the textured 3D assets with the highest quality.
Even for the text in the image prompt, our model can produce the correct bumps on the shape surface and an accurate texture map according to the geometric conditions.
The rest of the cases demonstrate the ability of our model to generate high-resolution and high-fidelity results with complex actions or scenes.

\begin{figure*}
\centering
\includegraphics[width=0.8\linewidth]{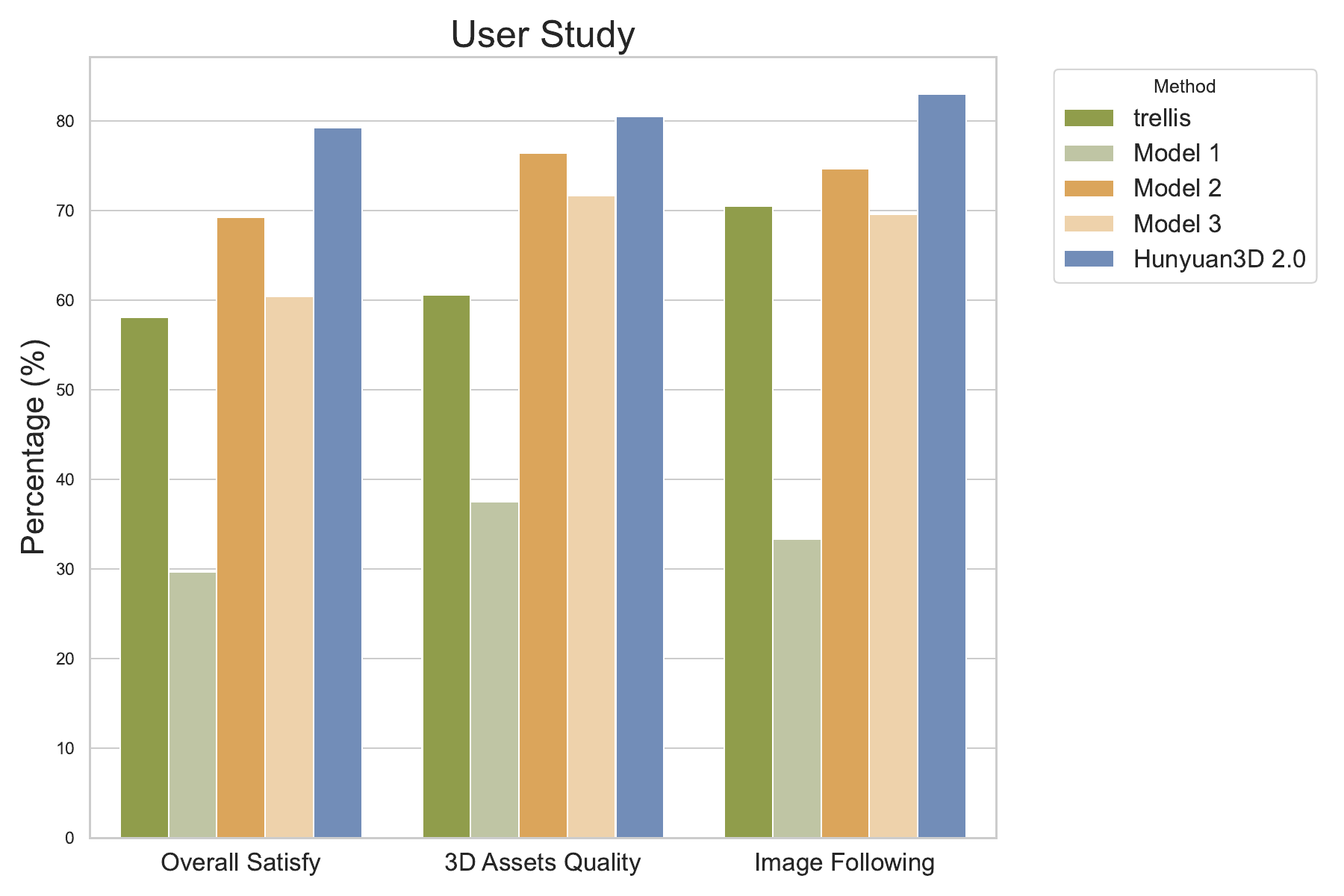}
\caption{The results of user study.}
\label{fig:user_study}
\end{figure*}

\textbf{User Study.}
In addition, we conducted a user study by randomly inviting 50 volunteers to evaluate 300 unselected results generated by Hunyuan3D 2.0 subjectively. The evaluation criteria included 1) overall visual quality, 2) adherence to image conditions, and 3) overall satisfaction (dissatisfaction in either 1 or 2 results in overall dissatisfaction).
The user study results in \cref{fig:user_study} indicate that Hunyuan3D 2.0 outperforms comparative methods, particularly in its ability to adhere to image conditions. 

\input{tables/end2end/end2end}

%% file: tables/end2end/end2end.tex
\begin{table}
\centering
\small
\begin{tabular}{rcccccccc}
\hline
            & CMMD($\downarrow$) & FID$_{CLIP}$($\downarrow$) & FID$_{Incept}$($\downarrow$) & CLIP-score($\uparrow$)    \\ \hline
Trellis~\cite{xiang2024structured}      & 3.591    & 54.639     & 289.287    & 0.787         \\ 
Model 1       & 3.600    & 55.866     & \underline{305.922}    & 0.779         \\ 
Model 2        & 3.368    & \underline{49.744}     & 294.628    & \underline{0.806}         \\ 
Model 3        & \underline{3.218}    & 51.574     & 295.691    & 0.799         \\ 
\shortname (Ours)  & \textbf{3.193}    & \textbf{49.165}     & \textbf{282.429}    & \textbf{0.809}           \\        \hline
\end{tabular}
\vspace{3mm}
\caption{Numerical comparison.
According to the results, Hunyuan3D-Paint produces the most condition-following texture maps.}
\label{tab:e2e}
\end{table}


%% file: sections/experiments/applications.tex

%% file: sections/applications.tex
\section{Hunyuan3D-Studio}
We have developed \href{https://arxiv.org/abs/2509.12815}{Hunyuan3D-Studio}~\cite{lei2025hunyuan3d}. This platform includes a comprehensive set of tools for the 3D production pipeline, as illustrated in ~\cref{fig:sys}.
Hunyuan3D-Studio aims to provide experts and novices with a no-frills way to engage in 3D generation production and research. 
In this section, we highlight several features of Hunyuan3D-Studio, including Sketch-to-3D, Low-polygon Stylization, and Autonomous Character Animator. These features aim to streamline the 3D creation process and make it accessible to a broader audience.

\subsection{Sketch-to-3D}

In game development and content creation, converting 2D sketches into 3D assets is a crucial technology that significantly enhances digital artistry design efficiency and flexibility.
Previous methods~\cite{bandyopadhyay2024doodle,zhou2023ga,zhang2021sketch2model,guillard2021sketch2mesh} suffer from the lack of the generative foundation model. They tackle this by training a small-scale generative or reconstruction model with sketch images as input directly on a limited dataset, significantly limiting the model's capabilities.

Benefitting from \shortname, we could convert sketches to images with rich details as input to the foundation 3D generative model. Specifically, the Hunyuan3D-Studio has developed the Sketch-to-3D module, which first converts sketches to images with rich details, maintaining original contours. Then, synthesize high-resolution and high-fidelity textured 3D assets, significantly reducing the barrier for users to engage in content creation.

As shown in \cref{fig:sys}, the Sketch-to-3D module can generate highly detailed and realistic 3D assets while maintaining close consistency with the original sketches. With this technology, users can synthesize 3D content with a simple sketch, providing a powerful tool for game developers and digital artists and a low-barrier creation platform for ordinary users. 

\subsection{Low-polygon Stylization}

Low-polygon stylization is critical in many computer graphics (CG) pipelines, as the face count of a mesh significantly impacts the application of 3D assets. Low-polygon stylization can significantly reduce computational costs, making it an essential process in 3D asset management.
To address this, we have established a low-polygon stylization module that efficiently converts the dense meshes generated by \shortname into low-polygon meshes. This module operates in two steps: geometric editing and texture preserving. 

For geometric editing, we employ a faster and more robust traditional method~\cite{garland1997surface,hoppe1999new}, despite the recent auto-regressive transformer-based polygon-generation approaches~\cite{weng2024scaling,xu2024cad,tang2024edgerunner,chen2024meshanything}. 
By setting an optimization criterion, we merge the vertices of the mesh to transform the dense mesh into a low-polygon mesh. As shown at the top-right of \cref{fig:sys}, each 3D model can be represented by only dozens of triangles after geometric editing.
The change in the face count of the mesh causes significant deviations in the vertices and faces of the low-polygon mesh compared to the dense mesh. 
Therefore, to preserve the texture patterns of the textured 3D assets, we construct a KD-tree for the input dense mesh. We then use the nearest-neighbor search within the KD tree to query the texture colors for the vertices of the low-polygon mesh.
Finally, we obtain the texture map for the low-polygon mesh by performing texture baking on the low-polygon mesh with vertex colors. This process ensures that the visual quality of the textures is maintained while optimizing the mesh structure for production-level textured 3D assets.

\subsection{3D Character Animation}
\shortname generates static 3D assets with high-resolution shapes and texture maps. 
However, drivable 3D models yet have broad requirements~\cite{starke2024categorical,starke2022deepphase,starke2021neural,starke2020local,liu2019liquid}, such as game development and animation production. 
To extend the range of applications of \shortname, we develop a 3D character animation function in Hunyuan3D-Studio. 
The animation algorithm inputs the generated character and extracts features from mesh vertices and edges.
Then, we utilize the Graph Neural Network (GNN) to detect skeleton key points and assign skinning weights to the mesh surface. 
Finally, based on the predicted skeleton skinning and motion templates, the algorithm utilizes motion retargeting to drive the character. 
Some frames are displayed in \cref{fig:sys}. 
With 3D character animation, the generated results from \shortname can come to life.

%% file: sections/related-work.tex
\section{Related Work}

\input{sections/related-work/geometry}

\input{sections/related-work/texture}

%% file: sections/related-work/geometry.tex
\subsection{3D Shape Generation}

\textbf{Representations.}
The field of shape generation has undergone significant advancements, driven by the unique challenges associated with the 3D modality. Unlike other modalities, 3D data lacks a universal storage and representation format, leading to diverse approaches in shape generation research. The primary 3D representations include voxels, point clouds, polygon meshes, and implicit functions. With the advent of seminal works such as 3D ShapeNets~\cite{wu20153d}, IF-Net~\cite{chen2019learning}, and 3D Gaussian Splatting~\cite{kerbl20233d}, implicit functions, and 3D Gaussian Splitting have become prevalent in shape generation. However, even lightweight and flexible representations like implicit functions impose substantial modeling and computational burdens on deep neural networks.
As a result, neural representations of 3D shapes have emerged as a new research focus, aiming to enhance efficiency. Pioneering methods such as 3DShape2VecSet~\cite{zhang20233dshape2vecset}, Michelangelo~\cite{zhao2024michelangelo}, CLAY~\cite{zhang2024clay}, and Dora~\cite{chen2024dora} represent 3D shapes using vector sets (one-dimensional latent token sequences proposed by 3DShape2VecSet~\cite{zhang20233dshape2vecset}), significantly improving representation efficiency. Another approach involves structured representations (e.g., triplane~\cite{peng2020convolutional,chan2022efficient,gao2022get3d} or sparse volume~\cite{mittal2022autosdf,zheng2023locally,ren2024xcube}) to encode 3D shapes, which better preserve spatial priors but are less efficient than vector sets. Inspired by recent advances in Latent Diffusion Models, \shortname employs vector sets to represent 3D shapes' implicit functions, alleviating the compression and fitting demands on neural networks and achieving a breakthrough in shape generation performance.

\textbf{Shape Generative models.}
The evolution of generative model paradigms has continually influenced shape generation research. Early works~\cite{wu2016learning,sanghi2022clip,yan2022shapeformer,yin2023shapegpt} based on Variational Auto-encoder~\cite{kingma2013auto}, generative adversarial networks (GANs)~\cite{goodfellow2014generative}, normalizing flow~\cite{papamakarios2021normalizing}, and auto-regressive modeling~\cite{gregor2014deep} demonstrated strong generative capabilities within several specific categories. The success of diffusion models~\cite{ho2020denoising} and their variants~\cite{liu2022flow,lipman2022flow} in text-conditioned image~\cite{rombach2022high,flux2024} generation has spurred the popularity of diffusion-based shape generative models, with notable works~\cite{cheng2023sdfusion,zhao2024michelangelo} achieving stable cross-category shape generation. Additionally, advancements in network architectures have propelled shape-generation research. The transition from early 3D Convolutional Neural Networks~\cite{graham2015sparse} to the now-common Transformer architectures~\cite{vaswani2017attention,peebles2023scalable} has led to the development of classic shape generation networks, enhancing performance. Building on these advancements, \shortname employs a flow-based scalable transformer, further improving the model's shape generation capabilities.

\textbf{Large-scale Dataset.}
Large-scale datasets are the cornerstone of scaling laws. However, the scale of 3D data is much smaller than that in large language models and image generation fields. From 3Dscanrep~\cite{curless1996volumetric,krishnamurthy1996fitting,turk1994zippered} to ShapeNet~\cite{chang2015shapenet}, the growth of 3D datasets has been gradual~\cite{stojanov2021using,zhou2016thingi10k,fu20213d,Selvaraju_2021_ICCV,xu2023animal3d,downs2022googlescannedobjectshighquality,collins2022abo}. The release of objaverse~\cite{objaverse} and objaverse-xl~\cite{objaverseXL} has been a significant driver in realizing the scaling law for shape generation. Leveraging these open-source 3D datasets, Hunyuan3D-2.0 can generate high-fidelity and high-resolution 3D assets.

Benefiting from these open-source algorithms and 3D datasets, \shortname is capable of generating high-fidelity and high-resolution 3D assets. Therefore, we have released \shortname to contribute to the open-source 3D generation community and further advanced 3D generation algorithms.

%% file: sections/related-work/texture.tex
\subsection{Texture Map Synthesis}
High-quality texture-map synthesis has been a long-standing topic within the computer graphics community. 
Its significance has only increased with the growing demand for end-to-end 3D generation techniques, where it plays a crucial role for appearance modeling.

\noindent\textbf{Text to Texture.}
Given a plain mesh, text/image to texture aims to generate a high-quality texture that aligns well with the given geometry according to a guided text and image.
Early attempts tried to approach texture synthesis by harnessing the categorical information and train a generative model on a specified dataset~\cite{chen2023shaddr,bokhovkin2023mesh2tex,dundar2023fine,siddiqui2022texturify,gao2022get3d,gao2021tm}. 
While achieving plausible texturing results, these methods failed to generalize to objects of other categories, limiting their applicability in production environments.

More recently, Stable Diffusion~\cite{rombach2022high}, owing to its impressive text-guided image generation capability and flexible structure, has spawned a plethora of text-to-texture research. To take full advantage of pre-trained image diffusion models, most subsequent works have approached the texture synthesis problem as a geometry-conditioned multi-view images generation problem.

Initially, score distillation was adopted to harness the generation power of image diffusion models for 3D content (texture) synthesis~\cite{tang2023dreamgaussian,lin2023magic3d,metzer2023latent,poole2022dreamfusion}. However, these methods are often limited by the over-saturated colors and misalignment with geometry.

Subsequently, optimization-free approaches pioneered by TEXTure~\cite{richardson2023texture} have been introduced~\cite{xiang2024make, cheng2024mvpaint, liu2025vcd, zhang2024texpainter, zeng2024paint3d, liu2024text, chen2023text2tex}. To ensure consistency across multi-view images, these methods either adopt an inpainting framework by specifying viewpoint-related masks or employ a "synchronizing" operation during the denoising process. However, since Stable Diffusion is trained on a dataset with a noticeable forward-facing viewpoint bias~\cite{liu2023zero}, these training-free methods are limited and often suffer from severe performance issues, such as the Janus problem and multi-view inconsistency, which result in textures with significant artifacts.

With the development of extensive 3D datasets, training multi-view diffusion models has become a prevailing direction for texture generation~\cite{lu2024genesistex2, bensadoun2024meta}, exhibiting more powerful capabilities on texture consistency than the training-free approaches.

\noindent\textbf{Image to Texture.}
In a related direction, image-guided texture generation has garnered attention in recent months, aligning closely with our research focus. This relatively unexplored area of image-guided texture synthesis demonstrates significant potential for further development since images provide more diverse information than text prompts, and text-to-texture generation can be fully replaced by a text-to-image and image-to-texture pipeline. Unfortunately, most of the existing works focus on semantic alignment with the reference image rather than precise alignment. 
FlexiTex~\cite{jiang2024flexitex} and EASI-Tex~\cite{sairaj_sig24} both utilize an IP-Adapter~\cite{ye2023ip-adapter} for image prompt injection. 
While TextureDreamer~\cite{yeh2024texturedreamer} employs a DreamBooth-like~\cite{ruiz2023dreambooth} approach to facilitate texture transfer across different objects.

However, we argue that there are two explicit advantages to exactly following every detail of the reference image. 
First, as part of an end-to-end image-guided 3D generation process, the geometry generated in the first stage strives to align with the reference image, while the appearance details are left for texture synthesis stage. 
Thus, one of the main objectives of our texture generation framework is to enhance the geometry with more detailed appearance features from the well-aligned reference image.
Second, with the rapid development of image diffusion techniques, more exquisite reference images are now available. Carefully adhering to these details can significantly improve the quality of the generated textures.
Based on these advantages,  Hunyuan3D-Paint is designed with a detailed preserving image injection module according to the philosophy of aligning the reference image not only semantically but also following the details as closely as possible.

\noindent\textbf{Multi-view Images Generation.}
Due to the viewpoint bias and multi-view inconsistency inherent in training-free image diffusion models, multi-view image diffusion was developed to alleviate these issues by utilizing large-scale 3D datasets, such as objaverse and objaverse-xl~\cite{objaverse, objaverseXL}.

Most works force the multi-view generated latents to communicate with each other by manipulating the self-attention layers with 3D-aware masks~\cite{huang2024mvadapter, li2024era3d, tang2025mvdiffusion++, long2024wonder3d, wang2023imagedream, shimvdream, shi2023zero123++, liu2023zero}. 
For example, Zero123++~\cite{shi2023zero123++} first treats the multi-view attention as a self-attention on a large image, which is the spatial concatenation of six multi-view images.
MVDiffusion~\cite{Tang2023mvdiffusion} applies a correspondence-aware attention (CAA) to inform the model to focus only on the correlation among the spatially-close pixels. 
MVAdapter~\cite{huang2024mvadapter}, following Era3D~\cite{li2024era3d} implements a simpler but effective row-wise and column-wise attention to alleviate the computational burden of CAA and achieves comparable performance.

Inspired by these works, we propose a multi-view generation framework equipped with a multi-task attention mechanism to achieve both multi-view consistency and image alignment simultaneously. 
Benefiting from this careful design and being trained on a large 3D rendering dataset, Hunyuan3D-Paint is able to achieve high-quality, consistent textures with strong alignment to the reference image.

%% file: sections/conclusion.tex
\section{Conclusion}

In this report, we introduce an open-source 3D creation system—\shortname—for generating textured meshes from images. We present Hunyuan3D-ShapeVAE, which is trained using a novel importance sampling method. This approach compresses each 3D object into a few latent tokens while minimizing reconstruction losses. Building on our VAE, we developed Hunyuan3D-DiT, an advanced diffusion transformer capable of generating visually appealing shapes that align precisely with input images. Besides, we introduce Hunyuan3D-Paint, another diffusion model designed to create textures for both our generated meshes and user-crafted meshes. With several innovative designs, our texture generation model, in conjunction with our shape generation model, can produce high-resolution, high-fidelity textured 3D assets from a single image. As we continue to make progress, we hope that \shortname will serve as a robust baseline for large-scale 3D foundation models within the open-source community and facilitate future research endeavors.

\clearpage

\section{Contributors}

\input{sections/contributors}

%% file: sections/contributors.tex
\label{contributors}
\begin{itemize}[leftmargin=0.25cm]
  \item \textbf{Project Sponsors:} Jie Jiang, Yuhong Liu, Di Wang, Yong Yang, Tian Liu
    \item \textbf{Project Leaders:} Chunchao Guo, Jingwei Huang, Zibo Zhao
    \item \textbf{Core Contributors:}
    \begin{itemize}[leftmargin=0.5cm]
        \item \textbf{Data:} Lifu Wang, Jihong Zhang, Meng Chen, Liang Dong, Yiwen Jia, Yulin Cai, Jiaao Yu, Yixuan Tang, Hao Zhang, Zheng Ye, Peng He, Runzhou Wu, Chao Zhang, Yonghao Tan
        \item \textbf{Shape Generation:} Zeqiang Lai, Qingxiang Lin, Yunfei Zhao, Haolin Liu
        \item \textbf{Texture Synthesis:} Shuhui Yang, Yifei Feng, Mingxin Yang, Sheng Zhang
        \item \textbf{Downstream Tasks:} Xianghui Yang, Huiwen Shi, Sicong Liu, Junta Wu, Yihang Lian, Fan Yang, Ruining Tang, Zebin He, Xinzhou Wang, Jian Liu, Xuhui Zuo, Song Zhang
        \item \textbf{Studio:}  Zhuo Chen, Biwen Lei, Haohan Weng, Jing Xu, Yiling Zhu, Xinhai Liu, Lixin Xu, Shaoxiong Yang, Yang Liu, Changrong Hu, Tianyu Huang, Shaoxiong Yang, Song Zhang, Yang Liu
    \end{itemize}
    \item \textbf{Contributors:} Jie Xiao, Yangyu Tao, Jianchen Zhu, Jinbao Xue, Kai Liu, Chongqing Zhao, Xinming Wu, Zhichao Hu, Lei Qin, Jianbing Peng, Zhan Li, Minghui Chen, Xipeng Zhang, Lin Niu, Paige Wang, Yingkai Wang, Haozhao Kuang, Zhongyi Fan, Xu Zheng, Weihao Zhuang, YingPing He
\end{itemize}

